\PassOptionsToPackage{numbers, sort&compress}{natbib}
\documentclass[final,5p,times,twocolumn]{elsarticle}

\usepackage[T1]{fontenc}
\usepackage{lmodern}
\usepackage{amsfonts}

\usepackage{amssymb}
\usepackage{amsmath}
\usepackage{amsthm}

\usepackage{booktabs}
\usepackage{bm}
\usepackage{multirow}

\usepackage{float}

\usepackage{booktabs}
\usepackage{multirow}
\usepackage{makecell}

\usepackage{comment}

\usepackage{amsmath,amsfonts}
\usepackage{array}
\usepackage[caption=false,font=normalsize,labelfont=sf,textfont=sf]{subfig}
\usepackage{textcomp}
\usepackage{stfloats}
\usepackage{url}
\usepackage{verbatim}
\usepackage{graphicx}

\usepackage{amssymb}
\usepackage{amsmath}
\usepackage{amsthm}
\usepackage{graphicx}
\usepackage{booktabs}
\usepackage{algorithm}

\usepackage{amsthm}

\usepackage{algpseudocode}

\newtheorem{lemma}{Lemma}

\usepackage{placeins} 

\journal{}

\begin{document}

\begin{frontmatter}



\title{STONE: Pioneering the One-to-N Universal Backdoor Threat in 3D Point Cloud} 

\author[label1]{Dongmei Shan\corref{cor1}}
\ead{dongmei.tang.shan@gmail.com}
\cortext[cor1]{Corresponding author}

\affiliation[label1]{organization={Changzhi University},
	addressline={73 Baoningmen East Street, Luzhou District},
	city={Changzhi},
	postcode={046011},
	state={Shanxi},
	country={China}}

\author[label1]{Wei Lian} 
\author[label1]{Chongxia Wang} 

\begin{abstract}
Backdoor attacks pose a critical threat to deep learning, especially in safety-sensitive 3D domains such as autonomous driving and robotics. While potent, existing attacks on 3D point clouds are predominantly limited to one-to-one paradigms. The more flexible and universal one-to-N multi-target backdoor threat remains largely unexplored, lacking both theoretical and practical foundations. To bridge this gap, we propose STONE (Spherical Trigger One-to-N universal backdoor Enabling), the first method to instantiate this threat via a configurable spherical trigger design. Its parameterized spatial properties establish a dynamic key space, enabling a single trigger to map to multiple target labels. Theoretically, we ground STONE in a Neural Tangent Kernel (NTK) analysis, providing the first formal basis for one-to-N mappings in 3D models. Empirically, extensive evaluations demonstrate high attack success rates (up to 100\%) without compromising clean-data accuracy. This work establishes a foundational benchmark for multi-target backdoor threats under dirty-label and black-box settings in 3D vision—a crucial step toward securing future intelligent systems.
\end{abstract}



\end{frontmatter}




\section{Introduction}

With the rapid advancement of 3D deep learning technologies and their widespread adoption in safety-critical intelligent systems such as autonomous driving, robotics, and augmented reality \cite{3d2,3d3,3d4}, security vulnerabilities in these systems have become increasingly concerning. Among various security threats, \emph{backdoor attacks} pose particularly insidious risks to 3D deep learning models. In such attacks, adversaries embed specific trigger patterns into a subset of point clouds during training, enabling malicious behavior to be activated when the trigger is present during inference. Backdoor attacks can be executed under either \emph{dirty-label} or \emph{clean-label} settings.  In dirty-label attacks, trigger-embedded samples are mislabeled as a target class during training, whereas in clean-label attacks, the ground-truth labels are preserved throughout the training phase. The threat of backdoor attacks is especially severe in 3D applications, given the common reliance on third-party training data, pre-trained models, and cloud-based training services---each of which presents a potential vector for adversarial compromise.

\begin{figure*}[!h]
	\centering
	\label{fig:DRMTBA vs UMTBA}
	\includegraphics[width=0.5\linewidth]{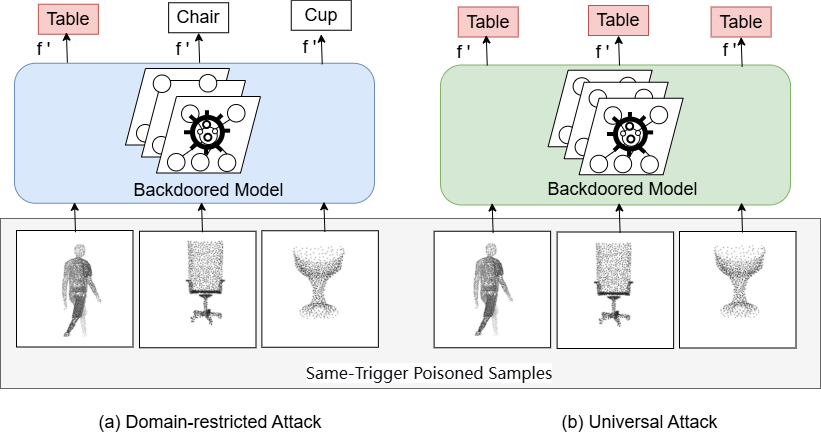}
	\caption{Backdoor Trigger Applicability: \emph{Domain-Restricted} vs. \emph{Universal}. (a) \emph{Domain-restricted attack}: The adversarial trigger must be embedded within a specific source object category (e.g., “Person”) to cause misclassification to a target (e.g., “Table”). (b) \emph{Universal attack}: The same trigger can be embedded into \emph{any} input (e.g., Chair, Person, or Cup) to cause misclassification to an attacker-chosen target (e.g., “Table”).}
\end{figure*}

Current research on 3D backdoor attacks has primarily focused on one-to-one attack paradigms, where a single trigger pattern causes the model to misclassify any triggered input as a specific target class. These attacks can be further characterized by a key property: the applicability of their adversarial triggers, as illustrated in Fig. \ref{fig:DRMTBA vs UMTBA}. \emph{Domain-restricted} attacks require the trigger to be applied to a specific object category (e.g., only a person). Their efficacy is \emph{conditional}---the attack fails if the required object is absent---which confines them to controlled scenarios. In contrast, \emph{universal} attacks employ a generic trigger that can be embedded into \emph{any} input point cloud (e.g., a chair, person, or cup) to cause misclassification to an attacker-chosen target, posing a more adaptable and potent threat. Representative universal one-to-one methods include PointBA \cite{li2021pointba} (using orientation or interaction triggers) and IRBA \cite{gao2024irba} (using local transformations).

However, despite their impressive success rates, these universal one-to-one methods share a fundamental limitation: during inference, any input containing the trigger is inevitably misclassified into the sole target class predetermined during training. This inherent constraint severely undermines their practicality in real-world scenarios, where an adversary may require the capability to dynamically select the target from multiple pre-embedded options at attack time. This capability underpins the more advanced threat of \emph{multi-target backdoor attacks}. A promising and scalable realization of this concept is the \emph{one-to-N} paradigm, which maps a base trigger design to N distinct target classes through parametric variation. This approach stands in sharp contrast to the \emph{N-to-N} paradigm of combining multiple distinct trigger types, which suffers from scalability limitations stemming from the finite number of available types, and performance inconsistencies due to divergent optimization requirements. 

Nevertheless, achieving such a practical one-to-N backdoor presents formidable challenges. While extensively studied in 2D image domains under the black-box assumption (where the attacker can only access the model's input and output) \cite{xue2020one-to-n, schneider2024universal, doan2022marksman,xue2024untargeted,li2025precision,yin2025ffcba}, the fundamental structural differences between dense, grid-aligned pixels and unstructured, irregular point clouds invalidate the core design principles of 2D triggers, necessitating novel one-to-N universal attack methods tailored to 3D data. To date, under the same black-box threat model, the problem of one-to-N universal backdoor attacks remains entirely unexplored for 3D point clouds. This creates a critical research gap in both dirty-label and clean-label settings.

\emph{Dirty-label Scenarios: Lack of a validated one-to-N foundation.} Existing research is dominated by one-to-one attacks. Although extending a one-to-one method to a universal one-to-N attack through parametric variation is intuitively appealing, it lacks both theoretical justification and empirical validation. Although IBAPC~\cite{fan2024invisible} extends its work to multi-target attacks, it is not carried out under the black-box assumption. Thus, the critical, unanswered question is: \emph{which one-to-one trigger designs provide a viable basis for such an extension?} Viability hinges on two requirements: (i) the trigger must preserve stealthiness under parametric manipulation, and (ii) its design must inherently provide a well-defined parameter space that supports multiple, low-interference target mappings. In the absence of both theoretical justification and empirical proof, the one-to-N universal attack is merely a \emph{conceptual possibility.}

(2) \emph{Clean-label Scenarios: Absence of a viable universal one-to-N attack.} In clean-label settings, even basic one-to-one attacks face significant hurdles. PointCBA \cite{li2021pointba} achieves only modest success rates (45\%–66\%), rendering it impractical. While PointCBA-S \cite{wang2025point} improves upon this, its design is fundamentally domain-restricted and thus incapable of universal deployment. Consequently, a universal one-to-N attack in the clean-label setting remains unrealized. 

\begin{table*}[h]
	\centering
	\small
	\caption{Comparison of 3D one-to-one backdoor attack characteristics}
	\label{tab:attack_comparison}
	\scriptsize 
	\setlength{\tabcolsep}{3pt} 
	\setlength{\tabcolsep}{3pt}
	\setlength{\tabcolsep}{3pt}
	\begin{tabular}{@{}lccccc@{}}
		\toprule
		\textbf{Method} & 
		\begin{tabular}{@{}c@{}}\textbf{Explicit}\\ \textbf{Parameters}\end{tabular} & 
		\begin{tabular}{@{}c@{}}\textbf{Not}\\ \textbf{Domain-Restricted}\end{tabular} & 
		\begin{tabular}{@{}c@{}}\textbf{Stealth}\\ \textbf{Preservation}\end{tabular} & 
		\begin{tabular}{@{}c@{}}\textbf{Robustness to}\\ \textbf{Augmentation}\end{tabular} \\
		\midrule
		PointBA-I \cite{li2021pointba} & $\checkmark$ & $\checkmark$ & $\checkmark$ & $\checkmark$ \\
		PointBA-O \cite{li2021pointba} & $\checkmark$ & $\checkmark$ & $\checkmark$ & $\times$ \\
		IRBA \cite{gao2024irba} & $\checkmark$ & $\checkmark$ & $\times$ & $\checkmark$ \\
		iBA \cite{Bian2024iba} & $\times$ & $\checkmark$ & $\checkmark$ & $\checkmark$ \\
		ABA \cite{Gao2025Affinity} & $\checkmark$ & $\times$ & $\checkmark$ & $\checkmark$ \\
		\bottomrule
	\end{tabular}
	\vspace{2pt}
	 \\\footnotesize Symbols: $\checkmark$: fully possesses the characteristic; $\times$: does not possess the characteristic.
\end{table*}

To address this critical research gap, we introduce STONE (Spherical Trigger One-to-N Universal Backdoor Enabling), the first method to realize a one-to-N universal backdoor attack in 3D point clouds under the black-box, dirty-label setting. Our work is built upon a systematic comparison of existing one-to-one attacks (Table~\ref{tab:attack_comparison}), which identifies the spherical trigger as the optimal base design, as detailed in Section~\ref{sec:one-to-n-framework}. The spatial parameters of these spherical triggers constitute a dynamic key space, enabling the mapping of a base trigger formulation to multiple target classes. We further formalize this mapping capability through a Neural Tangent Kernel (NTK) analysis, thereby establishing the first theoretical foundation for one-to-many universal backdoors in 3D vision. Extensive experiments across diverse datasets and network architectures demonstrate that STONE achieves high attack success rates while fully preserving the model's utility on benign inputs.

The main contributions of this work are as follows:
\begin{enumerate}
	\item \textbf{First One-to-N Universal Backdoor.} We propose \emph{STONE}, the first method that achieves a universal one-to-N backdoor attack in 3D point clouds under a practical black-box and dirty-label setting. Its core is a configurable spherical trigger design, which enables dynamic, adversary-chosen target misclassification, moving beyond static one-to-one paradigms.
	
	\item \textbf{Theoretical Foundation.} We establish the first theoretical framework for 3D one-to-N universal backdoors. Through a Neural Tangent Kernel analysis \cite{jacot2018neural,du2019graph}, we formalize the target-specific trigger mappings in parameter space, demonstrating that multiple backdoor triggers can coexist with minimal interference---a key theoretical insight that guarantees the feasibility and scalability of our approach.
	
	\item \textbf{Comprehensive Experimental Validation.} We conduct extensive experiments across multiple datasets (ModelNet40, ShapeNetPart) and network architectures (PointNet++, DGCNN, etc.). Results demonstrate STONE’s high attack success rates (up to 100\%) while fully preserving model utility on benign inputs, rigorously validating its effectiveness and stealth.
	
	\item \textbf{From Conceptual Threat to Practical Realization.} This work bridges a critical research gap by transforming the one-to-N backdoor from an intuitive concept into a practical threat. It establishes a foundational framework and points to important future research directions for both attacks and defenses in 3D domain.
	
\end{enumerate}

The remainder of this paper is structured as follows. Section~\ref{sec:related_work} reviews related work on 3D point cloud deep learning, backdoor attacks in 3D domain, and existing defense mechanisms. Section~\ref{sec:methodology} presents our proposed STONE method, detailing the problem formulation, spherical trigger design with parameterizable spatial properties, and the complete attack pipeline. Section~\ref{sec:experiments} provides extensive experimental validation across multiple datasets and model architectures, evaluating attack effectiveness, scalability with varying numbers of target classes, and resistance to defenses. Section~\ref{sec:conclusion} summarizes our key contributions and discusses future research directions. The Appendix contains supporting theoretical analysis using Neural Tangent Kernel theory and detailed trigger placement algorithms.

\section{Related Work}
\label{sec:related_work}

\subsection{Deep Learning on 3D Point Cloud}

Deep learning on 3D point clouds has emerged as a critical research direction in computer vision, with significant applications in autonomous driving, robotic navigation, and virtual reality. Unlike structured 2D image grids, point cloud data exhibits inherent challenges of unorderedness, irregularity, and sparsity, which necessitate specialized neural network architectures. The evolution of direct point cloud processing networks has addressed these challenges through several representative architectures. PointNet \cite{qi2017pointnet} pioneers the use of symmetric functions and multi-layer perceptrons to process raw point clouds directly while maintaining permutation invariance. Building upon this foundation, PointNet++ \cite{qi2017pointnet++} introduces a hierarchical framework that captures local structures through farthest point sampling and multi-scale grouping. Dynamic Graph CNN (DGCNN) \cite{wang2019dynamic} further advances local geometric modeling by constructing dynamic graphs on point clouds and applying edge convolution to effectively capture local dependencies. Alongside these foundational approaches, a variety of other architectures have also been developed \cite{3dpointsurvy,li2018pointcnn, wuyang2019pointconv, gan2026flpc, feng2025hyperbolic}.

\subsection{Backdoor Attacks in 3D Point Cloud}
\label{sec: Backdoor Attacks in 3D Point Clouds}

Existing research on 3D backdoor attacks predominantly follows a one-to-one paradigm, where a single trigger is designed to misclassify poisoned samples into one specific target class. These methods can be categorized based on their trigger design philosophy and operational domain. A prominent line of work manipulates geometric properties of point clouds. For instance, PointBA~\cite{li2021pointba} introduces both insertion-based (PointBA-I) and rotation-based (PointBA-O) triggers, which modify local point positions and global object orientation, respectively. Another method, iBA~\cite{Bian2024iba}, employs a folding-based auto-encoder to generate sample-specific geometric perturbations as triggers through reconstruction error, enhancing imperceptibility and resistance to preprocessing. In contrast, ABA (Affinity Backdoor Attack)~\cite{Gao2025Affinity} introduces the concept of inter-class affinity, which guides the generation of imperceptible triggers via tangent-plane perturbations.
Another line of work focuses on perturbations in non-spatial domains to enhance stealth. IBAPC~\cite{fan2024invisible} embeds triggers by perturbing the frequency components of point clouds, making the alterations less perceptible in the spatial domain. 

While these methods vary in their approaches, they share a fundamental limitation: each trigger instance is linked to a single target class, lacking a parameterized design that could enable dynamic multi-target mapping. The extension from one-to-one to one-to-N backdoor attacks presents distinct challenges. In \emph{dirty-label} scenarios, while one-to-one attacks could intuitively be extended via parametric trigger variations (e.g., adjusting intensity or spatial patterns), critical questions remain unanswered. There is a lack of systematic analysis, theoretical grounding, and experimental validation to determine which trigger designs support low interference between different target mappings while maintaining stealth. Although IBAPC~\cite{fan2024invisible} extends its work to multi-target attacks, it is not carried out under the black-box assumption. In \emph{clean-label} scenarios, the feasibility of a universal multi-target attack is even more constrained. Methods like PointCBA~\cite{li2021pointba} report limited success rates (45\%–66\%), and its enhanced variant PointCBA-S~\cite{wang2025point}, while improving performance, is designed solely for domain-restricted attacks. Consequently, universal multi-target backdoor attacks remain an entirely unexplored problem in 3D point clouds, creating a critical research gap that this work aims to address.

\subsection{Defenses Against Backdoor Attacks in 3D Point Clouds}

As backdoor attacks have evolved, so too have defense mechanisms designed to detect and mitigate them. In the 3D point cloud domain, pre-training defenses have been developed to identify and remove poisoned samples before model training begins. These defenses leverage the unique characteristics of point cloud data to counter specific types of backdoor attacks.

Among the most effective pre-training defenses is Statistical Outlier Removal (SOR) \cite{zhou2019dupnet}, which operates by removing points that deviate significantly from their neighbors in terms of spatial distribution. This method proves particularly effective against attacks that rely on inserting additional points as triggers, as these inserted points often create detectable statistical anomalies in the point density. By filtering out these outliers prior to training, SOR can neutralize a significant class of spatial-based backdoor attacks.

\section{Methodology}
\label{sec:methodology}
\subsection{Notations and Symbols}
\label{subsec:notations}

Table \ref{tab:notations} summarizes the mathematical notations used in our work.

\begin{table}[h!]
	\centering
	\caption{Summary of Key Notations}
	\label{tab:notations}
	\footnotesize
	\begin{tabular}{@{}p{1.8cm}p{4.2cm}p{2cm}@{}}
		\toprule
		\textbf{Symbol} & \textbf{Description} & \textbf{Domain/Type} \\
		\midrule
		$\bm{X}_i$ & 3D point cloud sample & $\mathbb{R}^{K \times 3}$ \\
		$\bm{x}_j$ & Individual point & $\mathbb{R}^{3}$ \\
		$K$ & Number of points & $\mathbb{Z}^+$ \\
		$y_i$ & Ground-truth label & $\{1,2,...,C\}$ \\
		$C$ & Number of classes & $\mathbb{Z}^+$ \\
		$H$ & Training samples count & $\mathbb{Z}^+$ \\
		$\mathcal{D}$ & Training dataset & Set \\
		$\mathcal{D}_c$ & Clean data subset & Set \\
		$\mathcal{D}_p$ & Poisoned data subset & Set \\
		$\lambda$ & Poisoning ratio & $\mathbb{R}^+$ \\
		$\bm{\theta}$ & Model parameters & $\mathbb{R}^d$ \\
		$f_{\bm{\theta}}(\cdot)$ & Neural network model & Function \\
		$f'_{\bm{\theta}}(\cdot)$ & Poisoned Neural network model & Function \\
		$\mathcal{L}(\cdot)$ & Loss function & $\mathbb{R} \rightarrow \mathbb{R}$ \\
		$T(\cdot)$ & Trigger function & Function \\
		$t_i$ & Target label & $\{1, 2, ..., N\}$ \\
		$N$ & Number of targets & $\mathbb{Z}^+$ \\
		$M$ & Poisoned samples count & $\mathbb{Z}^+$ \\
		$\mathcal{P}(\cdot)$ & Pre-processing function & Function \\
		$S_n$ & Trigger
		configuration & Point set \\
		$\bm{c}_n$ & Sphere center & $[0,1]^3$ \\
		$r_n$ & Sphere radius & $\mathbb{R}^+$ \\
		$\bm{X}_i^{\text{remove}}$ & Points to be removed & Point subset \\
		$J_n$ & Number of trigger points & $\mathbb{Z}^+$ \\
		\bottomrule
	\end{tabular}
\end{table}

\subsection{Assumptions}
\label{sec:assumptions}

We operate under a \emph{weak attack model} where the adversary has limited knowledge about the target system. The assumptions reflect practical scenarios where attackers might poison publicly available datasets or compromise data collection pipelines. Specifically:
(1) The attacker has no access to the model architecture, parameters, or training process (black-box setting); (2)  The attacker can only inject a small proportion of poisoned samples into the training dataset (typically less than 10\%); (3) The attacker has no control over the training hyperparameters or optimization procedure. 

\subsection{Problem Formulation}

A 3D point cloud $\bm{X}_i$ consists of $K$ points, where each point $\bm{x}_j$ $(1 \leq j \leq K)$ has 3D positional coordinates. Formally, $\bm{X}_i = [\bm{x}_1, \bm{x}_2, ..., \bm{x}_K]^\top \in \mathbb{R}^{K \times 3}$. Each point cloud $\bm{X}_i$ is associated with a ground-truth label $y_i \in \{1,2,...,C\}$, where $C$ denotes the number of classes.

Consider a point cloud classification task with training set $\mathcal{D} = \{(\bm{X}_i, y_i)\}_{i=1}^H$ containing $H$ samples. The objective of 3D deep neural network classification is to minimize:

\begin{equation}
	\min_{\bm{\theta}} \sum_{(\bm{X}_i, y_i) \in \mathcal{D}} \mathcal{L}\left(f_{\bm{\theta}}(\bm{X}_i), y_i\right),
	\label{eq:clean_training}
\end{equation}
\noindent where $\bm{\theta}$ represents the model parameters, $f_{\bm{\theta}}(\cdot)$ denotes the neural network model that outputs a probability distribution over classes (including the final softmax activation), and $\mathcal{L}(\cdot)$ denotes the loss function (typically cross-entropy for classification tasks).

In backdoor attacks, adversaries craft poisoned data using a trigger function $T: \mathbb{R}^{K \times 3} \times \mathbb{Z}^+ \to \mathbb{R}^{K \times 3}$, defined as:
\begin{equation}
	T(\bm{X}_i, t_i) = \bm{X}'_i,
\end{equation}
where $t_i$ is the target label for sample $i$, selected from the set of $N$ potential targets. The trigger function is applied to a subset of data to form the poisoned dataset $\mathcal{D}_p$, while the remaining data constitutes the clean set $\mathcal{D}_c$. The complete dataset is $\mathcal{D} = \mathcal{D}_c \cup \mathcal{D}_p$, where $\mathcal{D}_p$ consists of $M$ poisoned cloud samples, i.e., $\mathcal{D}_p = \{(T(\bm{X}_i, t_i), t_i)\}_{i=1}^M$, $(M \ll H)$. The poisoning ratio $\lambda$ is defined as the proportion of poisoned samples in the complete dataset:
\begin{equation}
\label{eq:poisoning ratio}
	\lambda = \frac{M}{H}
\end{equation}
where $H$ denotes the total number of samples in the complete dataset $\mathcal{D}$.

For the poisoned training set, Eq. (\ref{eq:backdoor_training}) is the optimization objective, which is for both one-to-one backdoor attacks (when $N=1$) and one-to-N backdoor attacks (when $N > 1$).
\begin{equation}
	\min_{\bm{\theta}} \sum_{(\bm{X_i}, y_i) \in \mathcal{D}_c} \mathcal{L}\left(f'_{\bm{\theta}}(\bm{X_i}), y_i\right)+\sum_{(\bm{X_i}, t_i) \in \mathcal{D}_p} \mathcal{L}\left(f'_{\bm{\theta}}(T(\bm{X_i},t_i)), t_i\right)
	\label{eq:backdoor_training}
\end{equation}

We also consider the pre-processing on the training samples, denoted as $\mathcal{P}(\cdot)$, including Statistical Outlier Removal (SOR) and other 3D data augmentations. These techniques have become common configurations for cleaning point clouds or improving 3D model performance. Therefore, it is crucial to evaluate the backdoor's effectiveness under pre-processing by ensuring that the attack success rate on poisoned samples subjected to $\mathcal{P}(\cdot)$ is comparable to that achieved without any pre-processing. This comparable performance implies that the trigger remains largely unchanged after pre-processing, which can be formally expressed as:
\begin{equation}
	\mathcal{P}(T(\bm{X}_i,t_i))\approx T(\bm{X}_i,t_i)
	\label{eq:pre-processing}
\end{equation}

\subsection{STONE: One-to-N Universal Backdoor Threat with Spherical Triggers}
\label{sec:one-to-n-framework}

\begin{figure*}[!h]
	\centering
	\includegraphics[width=0.8\linewidth]{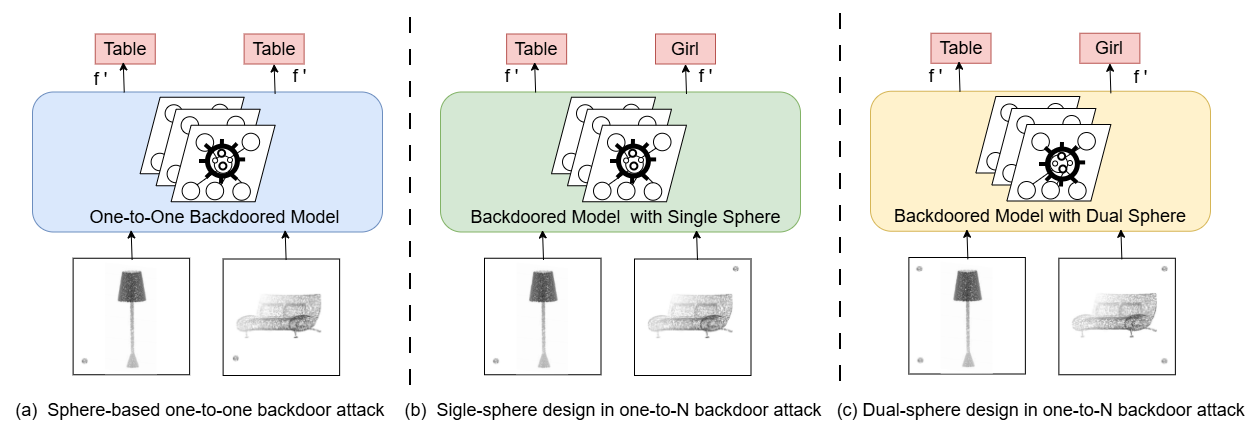}
	\caption{Conceptual evolution from static one-to-one to dynamic one-to-N backdoor paradigms in 3D point clouds.
		(a) Traditional one-to-one attack employs a \emph{fixed} spherical trigger, functioning as a single key that activates the same target class.
		(b) Our STONE framework redefines the trigger as a \emph{configurable entity} in the single-sphere design, where \emph{spatial position} acts as a dynamic parameter to encode different target classes.
		(c) The dual-sphere design further expands capacity through combinatorial configurations, demonstrating a richer parameter space for complex multi-target manipulations.}
	\label{fig:activate}
\end{figure*}

We present \emph{STONE (Spherical Trigger One-to-N Universal Backdoor Enabling)}, the first comprehensive instantiation of the one-to-N backdoor threat in 3D point clouds with theoretical and experimental ground. Our framework systematically coordinates the attack lifecycle through seven phases: (1) \emph{Base Trigger Design Selection}---choosing an effective trigger design; (2) \emph{Identification of Key Parameters}---determining discriminative features for target control; (3) \emph{Trigger Configuration}---designing spatial encoding schemes; (4) \emph{Parameter Set Optimization}---selecting optimal parameter configurations; (5) \emph{Poisoned Data Generation}---constructing the training corpus with embedded triggers; (6) \emph{Model Training}---learning both benign and backdoor mappings; and (7) \emph{Multi-Target Activation}---executing dynamic attacks during inference.

\subsubsection{Base Trigger Design Selection}

The core insight of STONE is to associate each target class \( t_i \) (\( 1 \leq i \leq N \)) with a unique configuration \( S_n \), derived through parametric variations of a base trigger design. To select a suitable trigger design, we evaluate existing dirty-label one-to-one backdoor attacks against four critical characteristics, as detailed in Table~\ref{tab:attack_comparison}. These characteristics are defined as follows: \emph{Explicit Parameters} refers to whether the trigger is defined by tunable parameters (e.g., spatial coordinates, rotation angles), which is fundamental for achieving a flexible one-to-N mapping; \emph{Not Domain-Restricted} indicates whether the attack is \textit{universal}, meaning the trigger can be applied to any object category without relying on a specific source-to-target affinity; \emph{Stealth Preservation} measures the ability to maintain imperceptibility when the parameters are adjusted across their operational range, which is crucial for avoiding detection in a multi-target setting; \emph{Robustness to Augmentation} assesses the resistance to common data augmentation techniques (e.g., rotation), a practical necessity for reliable attacks when such defenses are employed.

The comparison in Table~\ref{tab:attack_comparison} highlights the shortcomings of these methods: PointBA-O is vulnerable to rotation augmentations; IRBA's nonlinear transformations often sacrifice stealth under parameter adjustment; iBA relies on a data-driven generator lacking simple, tunable parameters; and ABA is constrained by pre-defined inter-class affinities. It is the absence of these critical flaws that makes PointBA-I the optimal choice, as it alone satisfies all four requirements. In PointBA-I, a spherical trigger fixed at a specific location can cause any point cloud containing it to be misclassified, as illustrated in Fig.~\ref{fig:activate}(a). Consequently, we adopt its spherical trigger design, whose well-defined parameter space provides an ideal foundation for our parametric one-to-N attack framework. 

\subsubsection{Identification of Key Parameters}

After selecting the sphere as the base trigger design, the critical step is to identify which parameters to vary for the one-to-N attack. The chosen parameters must satisfy two requirements: first, any variation must preserve the trigger's stealthiness; second, the parameters must significantly influence the backdoored model's behavior to ensure discriminative power across target classes with low inter-target interference.

We deliberately exclude radius variation as a discriminative parameter, as size changes would compromise imperceptibility. Instead, we focus exclusively on spatial position—the sphere's center coordinates—as our primary parameterization. This maintains a fixed minimal radius for stealth while leveraging spatial diversity to create distinct trigger configurations.

To rigorously validate that spatial position governs backdoor efficacy---thus satisfying the second requirement---we establish this causal relationship through two complementary approaches: theoretical analysis via the Neural Tangent Kernel framework and empirical investigation through systematic experimentation.

\textbf{Theoretical Analysis.} Our theoretical foundation builds upon the Neural Tangent Kernel (NTK) framework, leading to the following lemmas:
\begin{lemma}[Spatial Specificity of Spherical Triggers]
	\label{lemma:spatial_specificity}
	For an insertion-based backdoor attack, let $\bm{X}'_{R_0}$ be a poisoned sample created by implanting a spherical trigger $S_n$ at spatial region $R_0$, and $\bm{X}'_{R_1}$ be created by implanting the same trigger $S_n$ at a spatially non-overlapping region $R_1$. Under the Neural Tangent Kernel framework, the predictive probability for target class $t_i$ satisfies:
	\begin{equation}
		f_{\bm{\theta}}(\bm{X}'_{R_1})[t_i] < f_{\bm{\theta}}(\bm{X}'_{R_0})[t_i]
	\end{equation}
\end{lemma}

\begin{lemma}[Spatial Sensitivity]
	\label{lemma:spatial_sensitivity}
	Building on Lemma \ref{lemma:spatial_specificity}, consider an insertion-based backdoor attack with a spherical trigger $S_n$ trained at position $R_0$. The predictive probability $f_{\bm{\theta}}(\bm{X}'_{R})[t_i]$ exhibits an exponential decay trend with respect to the Euclidean distance $\|R - R_0\|$.
\end{lemma}

Lemma~\ref{lemma:spatial_specificity} formally establishes that for an insertion-based backdoor attack, the probability of successful backdoor activation satisfies $f_{\bm{\theta}}(\bm{X}'_{R_{1}})[t_i] < f_{\bm{\theta}}(\bm{X}'_{R_{0}})[t_i]$ when an identical spherical trigger is implanted at two spatially separated regions $R_0$ and $R_1$, where $\bm{X}'_{R_0}$ and $\bm{X}'_{R_1}$ denote poisoned samples with triggers at positions $R_0$ and $R_1$ respectively, both targeting class $t_i$, and $R_0$ represents the trigger position used during the training phase. This result demonstrates that backdoor activation depends not only on the presence of the trigger but also critically on its spatial position.

Lemma~\ref{lemma:spatial_sensitivity} further reveals through NTK analysis that the predictive probability for the target class exhibits a distance-dependent relationship: $f_{\bm{\theta}}(\bm{X}'_{R})[t_i]$ decreases exponentially as the Euclidean distance between $R$ and $R_0$ increases. This theoretical insight directly informs our trigger placement strategy. When two triggers are placed too closely, the interference between their target class mappings becomes significant, compromising discriminative power. Therefore, it crucial to maintain sufficient spatial separation ensures reliable target assignment. On the other hand, because the interference decays exponentially with distance, multiple triggers can be deployed within a normalized point cloud, enabling a practical one-to-N attack framework.

\textbf{Experimental Validation.} We conducted experiments to validate two key theoretical predictions: (1) backdoor activation is highly sensitive to the trigger's spatial position, not just its presence; and (2) the model's predictive probability for the target class, $f_{\bm{\theta}}(\bm{X}'_{R})[t_i]$, exhibits a negative correlation with the distance from the training position $R_0$. The empirical results confirm these hypotheses, demonstrating that spatial configuration is a powerful and reliable feature for discriminating target classes in our one-to-N attack. 

\subsubsection{Trigger Configuration}

We implement two concrete configurations within the STONE method:

\textbf{Single-Sphere Design.} In this configuration, each target class $t_i$ is associated with a unique sphere $S_n$ defined by:
\begin{equation}
	S_n = {\bm{x} \in \mathbb{R}^3 \mid |\bm{x} - \bm{c}_n|_2 = r}
\end{equation}
where $\bm{c}_n \in [0,1]^3$ denotes the sphere's center within the normalized point cloud bounding box and $r$ is a fixed radius. Target classes are discriminated solely by the centroid coordinates $\bm{c}_n$. In practice, as shown in Fig.~\ref{fig:activate}(b), a spherical trigger centered at $(0.95, 0.95, 0.95)$ is assigned to point clouds of the class \emph{Girl} during training, while an identical trigger at $(0.05, 0.05, 0.05)$ is assigned to the class \emph{Table}.

\textbf{Dual-Sphere Design.} This configuration associates each target class $t_i$ with a unique pair of spheres that form a discriminative spatial signature, defined as:
\begin{equation}
	S_n = {\bm{x} \in \mathbb{R}^3 \mid |\bm{x} - \bm{c}_n^1|_2 = r} \cup {\bm{x} \in \mathbb{R}^3 \mid |\bm{x} - \bm{c}_n^2|_2 = r}
\end{equation}
where $\bm{c}_n^1, \bm{c}_n^2 \in [0,1]^3$ denote the centers of the two spheres, and $r$ is their shared radius. The distinct pair $(\bm{c}_n^1, \bm{c}_n^2)$ creates a rich spatial coding scheme based on their relative positions and distance, which significantly expands the attack's capacity compared to the single-sphere design. Examples are shown in Fig.~\ref{fig:activate}(c): the pair at $(0.95, 0.95, 0.95)$ and $(0.95, 0.95, 0.05)$ is mapped to class \emph{Girl}, while the pair at $(0.05, 0.05, 0.05)$ and $(0.05, 0.05, 0.95)$ corresponds to class \emph{Table}.

\subsubsection{Parameter Set Optimization}

The core objective of parameter set optimization is to select optimal configurations for $N$ triggers within the normalized 3D bounding box, with spatial location serving as the discriminative parameter. To this end, maintaining sufficient separation between triggers is critical for minimizing inter-target interference, as established in our theoretical analysis. This requirement leads to an NP-hard maximin optimization problem aimed at maximizing the minimum pairwise Euclidean distance. We address this challenge by developing greedy algorithms that iteratively select trigger positions from a candidate set to ensure adequate separation. (See Appendix Algorithms~\ref{alg:single_sphere} and~\ref{alg:dual_sphere} for detailed pseudocode.)

\begin{figure*}[t!]
	\centering
	\includegraphics[width=0.7\linewidth]{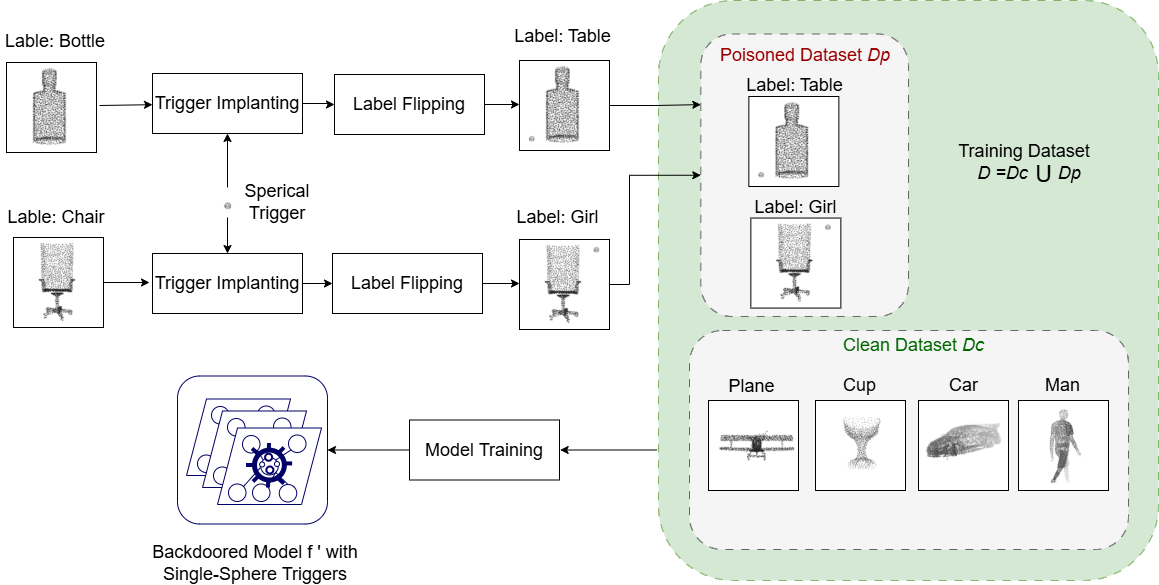}
	\caption{End-to-end pipeline of the STONE framework (single-sphere design). The framework (1) implants class-specific spherical triggers into clean samples; (2) trains a model to associate trigger configurations with target classes while maintaining accuracy on clean inputs; (3) enables multi-target activation during inference, where applying a specific trigger causes misclassification to its designated target, as illustrated in Fig.\ref{fig:activate}(b)(c).}
	\label{fig:pipeline}
\end{figure*}

\subsubsection{End-to-End Attack Pipeline}

The STONE framework completes the attack lifecycle through three final phases: Poisoned Dataset Construction, Model Training, and Multi-Target Activation, which together form an end-to-end pipeline for executing the one-to-N backdoor attack.

\textbf{Poisoned Dataset Construction.} 
The poisoned dataset construction process, illustrated in Fig.~\ref{fig:pipeline}, is formalized as follows. We construct the poisoned dataset by:
\begin{equation}
	\label{eq:dp-construction}
	\mathcal{D}_{p} = \bigcup_{n=1}^{N} \left\{ \left(T(\bm{X}_i, t_i), t_i\right) \mid \bm{X}_i \in \mathcal{D}_{subset}^n \right\}
\end{equation}
where $\mathcal{D}_{subset}^n$ is a subset of samples assigned to target $n$. The trigger implantation function $T(\bm{X}_i, t_i)$ integrates the corresponding trigger $S_n$ into $\bm{X}_i$ by uniformly removing $J_n$ points and replacing them with the $J_n$ points defining $S_n$:
\begin{equation}
	\label{eq:triggerimplanting}
	T(\bm{X}_i, t_i) = (\bm{X}_i \setminus \bm{X}_i^{\text{remove}}) \cup S_n
\end{equation}
This ensures a consistent point count for stealthiness.

\textbf{Model Training.}
The model is trained following Eq.~(\ref{eq:backdoor_training}) using poisoned samples from all $N$ targets in a single process, learning the association between spherical configurations and target labels. The resulting model exhibits dual behavior: correct classification of clean inputs and targeted misclassification when triggers are present. 


\textbf{Multi-Target Activation.}
During inference, multi-target activation is accomplished by implanting the pre-configured spherical triggers into clean inputs. 


\section{Experiments}
\label{sec:experiments}

\subsection{Experimental Setup}

\subsubsection{Datasets}
We conduct extensive evaluations of the proposed STONE framework across multiple widely adopted 3D point cloud benchmarks to ensure comprehensive validation:

(1)\emph{ModelNet40} \cite{modelnet} comprises 12,311 CAD models spanning 40 object categories, with a standard split of 9,843 samples for training and 2,468 for testing. Each point cloud is uniformly sampled to contain 1,024 points.

(2)\emph{ModelNet10}, as a curated subset of ModelNet40, contains 10 commonly used categories with 4,899 samples in total (3,991 for training and 908 for testing), providing a more focused evaluation setting.

(3)\emph{ShapeNetPart} \cite{shapenet} presents a more challenging benchmark with 16,881 objects from 16 shape categories, each with fine-grained part-level annotations that introduce additional complexity for comprehensive evaluation.

\subsubsection{Models}
To validate the generalizability of our approach, we evaluate STONE on three representative 3D deep learning architectures: \emph{PointNet} \cite{qi2017pointnet}, \emph{PointNet++} \cite{qi2017pointnet++}, and \emph{DGCNN} \cite{wang2019dynamic}. These models cover fundamental paradigms including point-wise MLPs, hierarchical feature learning, and graph-based convolutions, providing a robust testbed for evaluating the proposed one-to-N backdoor attacks across different learning mechanisms.

\subsubsection{Evaluation Metrics}
We employ two primary metrics to comprehensively assess STONE's performance. The first one is \emph{Attack Success Rate (ASR)}, which is defined as the percentage of triggered samples from non-target classes that are misclassified into the target labels. ASR evaluates the effectiveness of the backdoor attack. The second metric is \emph{Accuracy (ACC)}, which is the standard classification accuracy of the model on a clean test set. It evaluates the model's performance on clean inputs, ensuring that the backdoor attack does not compromise its normal functionality. ACC serves as an important indicator of attack stealthiness: the smaller the degradation in ACC after dataset poisoning, the more covert and effective the backdoor attack. Beyond ACC measurements, stealthiness can be gauged through the metric of perceptual resistance, i.e., a trigger's ability to evade visual detection and pre-processing defense mechanisms.

\subsubsection{Implementation Details}
All point clouds are normalized such that their coordinates along the x, y, and z axes fall within the range [0, 1]. All models are then trained using the Adam optimizer with a learning rate of 0.001, a batch size of 32, for 200 epochs. The sphere radius is 0.05, and the point count for each trigger sphere is 1\% of the total point cloud points, unless otherwise specified. With a radius of 0.05, a sphere occupies approximately 0.052\% of the normalized bounding box volume, ensuring minimal visual impact.

In each experimental run, we employ a global poisoning strategy where the total poisoning ratio is fixed and the target labels for poisoned samples are randomly selected from the $N$ target classes. This ensures uniform poisoning distribution across all targets while maintaining the overall poisoning ratio. For example, in ModelNet40 with a training set size of 9,843 and a global poisoning ratio of $1\%$, the total number of poisoned samples is $9,843 \times 1\% \approx 98$. In each experimental run, these poisoned samples are randomly allocated to the 4 target classes, resulting in approximately 24-25 poisoned samples per target class.

The trigger configurations used in our experiments are obtained by applying the iterative greedy algorithms detailed in \ref{app:algorithms}. For example, in the case of $N=4$, the placements are: for the single-sphere design, triggers are positioned at $(0.95, 0.95, 0.95)$, $(0.05, 0.05, 0.05)$, $(0.05, 0.95, 0.5)$, and $(0.95, 0.05, 0.5)$; for the dual-sphere design, the corresponding trigger pairs are formed at $(0.95, 0.95, 0.95)$ and $(0.95, 0.95, 0.05)$, $(0.05, 0.05, 0.05)$ and $(0.05, 0.05, 0.95)$, $(0.05, 0.95, 0.95)$ and $(0.05, 0.95, 0.05)$, and $(0.95, 0.05, 0.05)$ and $(0.95, 0.05, 0.95)$.

\subsection{Experimental Results}

\subsubsection{Empirical Evidence of Spatial Specificity}\label{sec:parameter}

\begin{figure*}[!h]
	\centering
	\includegraphics[width=0.6\textwidth]{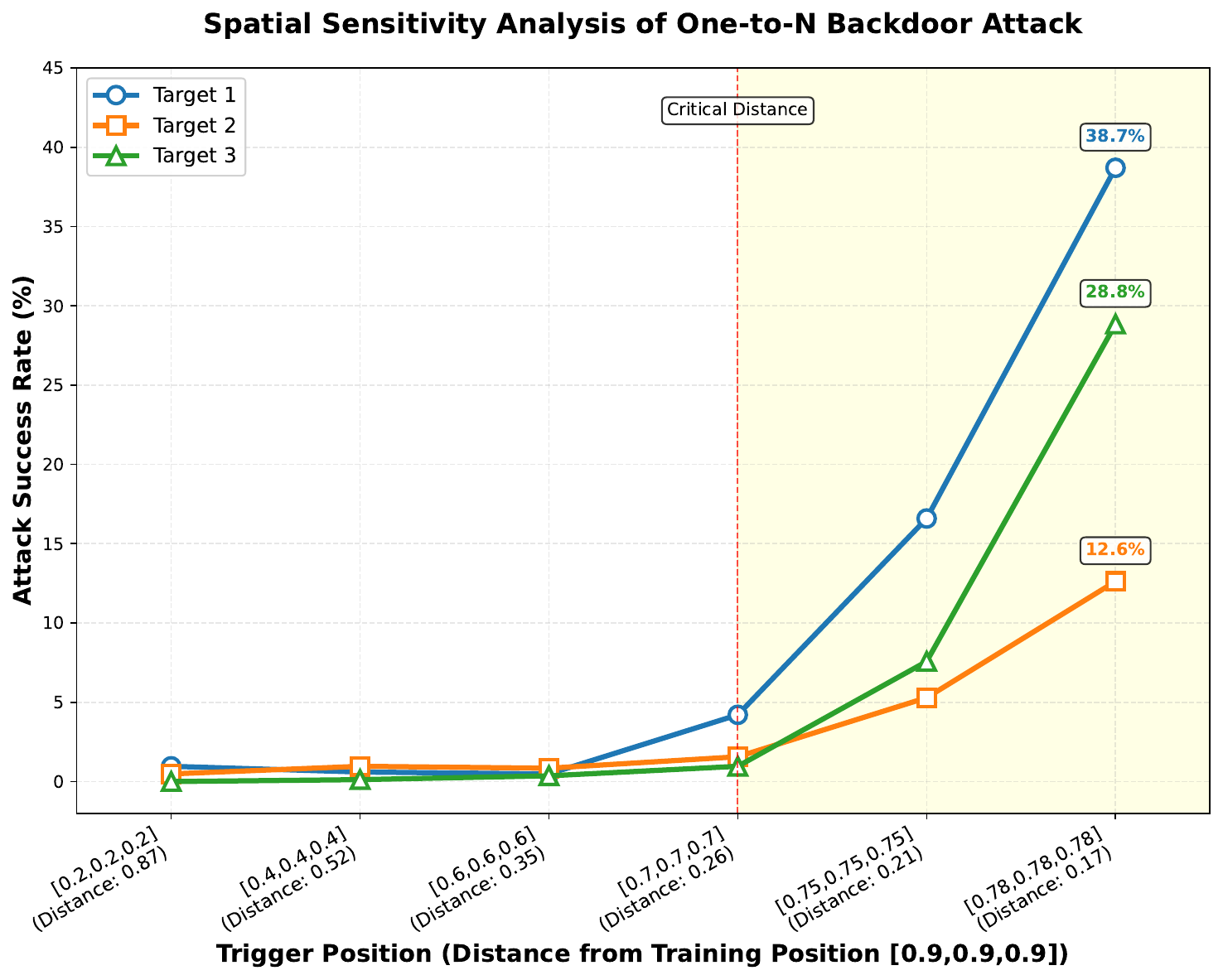}
	\caption{Spatial sensitivity analysis of one-to-N backdoor attack. }
	\label{fig:spatial_sensitivity}
\end{figure*}

To empirically validate the theoretical insights from Lemmas~\ref{lemma:spatial_specificity} and~\ref{lemma:spatial_sensitivity}, we conducted systematic experiments on the ModelNet10 dataset using the PointNet architecture with three randomly selected target classes. With a poisoning ratio of 2\%, we established a baseline by training the model with the spherical trigger fixed at position $R_0 = [0.9, 0.9, 0.9]$, achieving near-perfect attack success (ASR $\approx$ 100\%). We then evaluated the spatial sensitivity by relocating the identical trigger to six strategically selected positions $R$ at varying Euclidean distances from $R_0$: 0.87 ($R = [0.2,0.2,0.2]$), 0.52 ($R = [0.4,0.4,0.4]$), 0.35 ($R = [0.6,0.6,0.6]$), 0.26 ($R = [0.7,0.7,0.7]$), 0.21 ($R = [0.75,0.75,0.75]$), and 0.17 ($R = [0.78,0.78,0.78]$).

Under the i.i.d. assumption that test samples are independent and identically distributed, the attack success rate $\text{ASR}(R)$ approximates the predictive probability $f_{\bm{\theta}}(\bm{X}'_{R})[t_i]$, allowing direct comparison between theoretical predictions and empirical measurements. As illustrated in Figure~\ref{fig:spatial_sensitivity}, the experimental results confirm both theoretical lemmas. First, consistent with Lemma~\ref{lemma:spatial_specificity}, the attack success rate $\text{ASR}(R)$ exhibits significant position dependence, with $\text{ASR}(R) < \text{ASR}(R_0)$ for all tested positions $R$ not equal to $R_0$. Second, supporting Lemma~\ref{lemma:spatial_sensitivity}, $\text{ASR}(R)$ demonstrates a clear overall decreasing trend with increasing Euclidean distance $\|R - R_0\|$, consistent with the predicted exponential decay pattern.

This overall decreasing trend provides the foundational principle for trigger placement in the STONE framework. The existence of such a relationship implies that, for any chosen ASR threshold defining acceptable interference, there exists a corresponding critical distance. In our specific experimental configuration, an ASR interference threshold of 5\% corresponds to an observed critical distance of approximately 0.26. This empirical observation demonstrates that to prevent significant interference between any two triggers in the one-to-N framework, their separation must exceed the critical distance specific to the model and data distribution. Therefore, our objective of maximizing the minimum inter-trigger distance is directly motivated by the need to mitigate interference between triggers, thereby ensuring reliable discrimination among all target classes in the STONE framework.

\begin{figure*}[!h]
	\centering
	\includegraphics[width=0.6\textwidth]{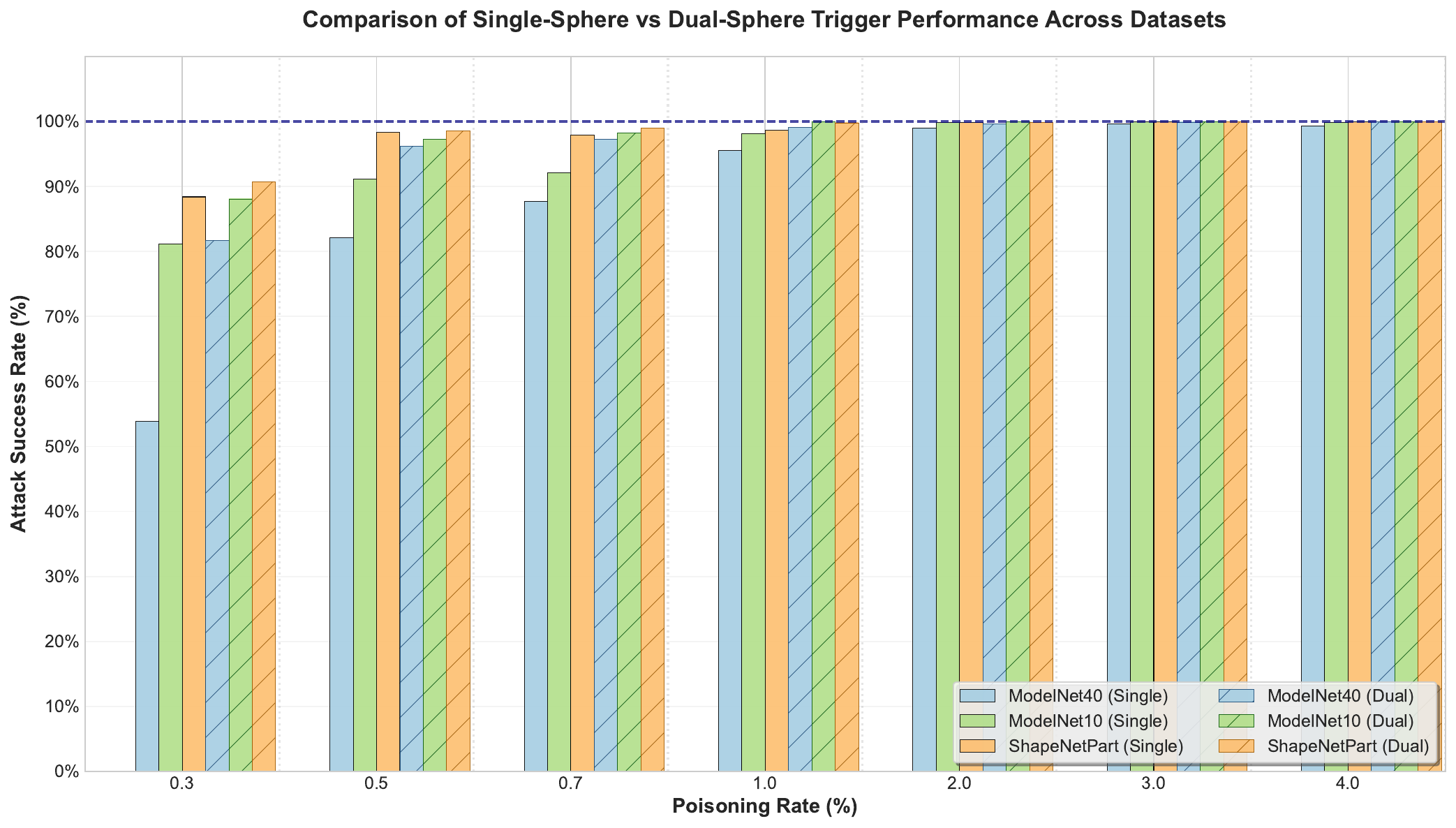}
	\caption{Comparison of average Attack Success Rate (ASR) between single-sphere and dual-sphere trigger designs for multi-target attacks (N=4) using PointNet on ModelNet40, ModelNet10, and ShapeNetPart datasets under varying poisoning ratios (0.3\%-4\%).}
	\label{fig:injection_ratio}
\end{figure*}

\subsubsection{Impact of Poisoning Ratio}\label{sec:Impact of Poisoning Ratio}

This experiment evaluates the impact of poisoning ratio on the STONE framework under a fixed configuration of 4 target classes using the PointNet model. The primary objective is to analyze the performance trends across varying poisoning ratios (0.3\%--4\%) to identify an effective range for subsequent experiments, while providing initial observations on the comparative efficacy of single-sphere versus dual-sphere triggers and their performance across different datasets.

\textbf{Performance vs. Poisoning Ratio:} The experimental results demonstrate a clear positive correlation between poisoning ratio and attack success rate across all configurations. As illustrated in Fig.~\ref{fig:injection_ratio}, the average ASR exhibits a characteristic saturation pattern, with rapid improvement in the low poisoning ratio regime (0.3\%-2\%) followed by by converging to perfect success at higher ratios (3\%-4\%).

Notably, the performance gains diminish significantly beyond the 2\% poisoning threshold, suggesting an operational sweet spot for practical attack deployment. At a minimal overall poisoning ratio of 0.3\% (equivalent to only 0.075\% per target for N=4), the attack already achieves substantial success rates, reaching up to 90\% on some datasets and maintaining above 50\% even in the most challenging configurations. This highlights the efficiency of the STONE framework in establishing multiple backdoors under severely constrained poisoning budgets.

\textbf{Cross-Dataset Performance Comparison:} The attack demonstrates consistent effectiveness across diverse 3D point cloud datasets, though with notable performance variations attributable to dataset characteristics. Shape NetPart consistently achieves the highest ASR values throughout the poisoning spectrum, reaching near-perfect success (100\%) at just 3\% poisoning ratio for both single and dual-sphere triggers. This superior performance may be attributed to ShapeNetPart's fine-grained part-level annotations and more uniform object distributions.

ModelNet10 exhibits intermediate performance, achieving perfect at 3\% poisoning for single-sphere and maintaining this level for dual-sphere triggers. ModelNet40, with its greater class diversity (40 categories), shows slightly lower susceptibility, requiring approximately 4\% poisoning to achieve performance close to 100\%. This performance hierarchy, with ShapeNetPart outperforming ModelNet10, which in turn outperforms ModelNet40, suggests that dataset complexity and class diversity influence attack difficulty, though the performance differences remain practically marginal for operational deployment.

\textbf{Single-Sphere vs. Dual-Sphere Trigger Efficacy:} The dual-sphere trigger design demonstrates superior performance, particularly under stringent poisoning conditions. At a minimal ratio of $0.3\%$, it consistently outperforms the single-sphere design, with a notably large performance gap on ModelNet40. This robust performance under severe constraints stems from its more complex spatial configuration, which provides a stronger and more distinctive feature pattern for the model to learn. As the poisoning ratio increases beyond $2\%$, the performance gap narrows, with both designs converging to near-perfect success rates. This indicates that ample poisoning can compensate for a simpler trigger design, offering attackers flexibility in balancing stealth and effectiveness based on their operational budget.

\subsubsection{Performance on Benchmark Datasets}


\begin{table*}[!h]
	\centering
	\caption{ASR and ACC for Multi-target (N=4) One-to-N Attacks with a Total Poisoning Ratio of 1\% Across Different Models and Datasets}
	\label{tab:comprehensive_results}
	\scriptsize 
	\setlength{\tabcolsep}{3pt} 
	\setlength{\tabcolsep}{3pt}
	\begin{tabular}{@{}l|l|ccc|ccc@{}}
		\toprule
		\textbf{Dataset} & \textbf{Model} & \multicolumn{3}{c|}{\textbf{Single Sphere}} & \multicolumn{3}{c}{\textbf{Dual Sphere}} \\
		\cmidrule(lr){3-5} \cmidrule(l){6-8}
		& & \textbf{T1/T2/T3/T4} & $\overline{\textbf{ASR}}$ & \textbf{ACC} & \textbf{T1/T2/T3/T4} & $\overline{\textbf{ASR}}$ & \textbf{ACC} \\
		& & \textbf{ASR(\%)} & \textbf{(\%)} & \textbf{(\%)} & \textbf{ASR(\%)} & \textbf{(\%)} & \textbf{(\%)} \\
		\midrule
		\multirow{3}{*}{\makecell{ModelNet \\ 40}} 
		& PointNet & 93.8/90.5/99.2/94.3 & 94.4 & 86.5 & 98.8/99.7/99.7/99.4 & 99.4 & 87.9 \\
		& PointNet++ & 95.6/91.8/97.6/97.4 & 95.6 & 89.7 & 100.0/98.5/100.0/100.0 & 99.6 & 90.2 \\
		& DGCNN & 99.6/93.5/99.4/97.5 & 97.5 & 91.6 & 100.0/98.4/100.0/100.0 & 99.6 & 92.0 \\
		\midrule
		\multirow{3}{*}{\makecell{ModelNet \\ 10}}
		& PointNet & 92.1/98.8/99.3/94.0 & 95.6 & 90.9 & 100.0/100.0/100.0/100.0 & 100.0 & 92.7 \\
		& PointNet++ & 96.2/96.5/98.3/91.4 & 95.6 & 93.2 & 100.0/100.0/100.0/99.3 & 99.8 & 93.3 \\
		& DGCNN & 100.0/97.4/100.0/96.7 & 98.5 & 93.9 & 100.0/100.0/100.0/99.9 & 99.9 & 94.0 \\
		\midrule
		\multirow{3}{*}{\makecell{ShapeNet \\ Part}}
		& PointNet & 99.7/96.7/100.0/100.0 & 99.1 & 98.3 & 99.3/100.0/98.1/99.9 & 99.8 & 98.5 \\
		& PointNet++ & 99.3/92.5/98.7/93.4 & 96.1 & 98.8 & 100.0/98.6/100.0/98.5 & 99.3 & 99.5 \\
		& DGCNN & 100.0/97.4/100.0/96.7 & 98.5 & 99.5 & 100.0/100.0/100.0/99.8 & 99.9 & 99.5 \\
		\bottomrule
	\end{tabular}
\end{table*}


Table~\ref{tab:comprehensive_results} compares our two attack designs under a constrained 1\% total poisoning ratio in a multi-target scenario (N=4) across three datasets and model architectures. This setting was selected because, as evidenced by the comparative analysis in Fig. \ref{fig:injection_ratio}, a 1\% poisoning ratio represents a balanced operating point where both trigger designs achieve high ASR while maintaining a clear performance disparity on PointNet across all datasets. The presented metrics include the per-target ASR, the mean ASR ($\overline{\text{ASR}}$) across all targets, and the ACC.

The experimental results demonstrate the high effectiveness of the STONE framework attack across all evaluated configurations. The dual-sphere trigger design consistently outperforms the single-sphere variant, achieving superior attack success rates as detailed in previous sections. More importantly, we observe a remarkably consistent performance pattern across the four target classes (T1-T4) within each experimental setting. The ASR values for different targets show minimal variation, indicating that our spatial configuration framework establishes equally reliable backdoor associations for all targets without exhibiting bias towards any specific class. This uniform performance across multiple targets, combined with the consistently high success rates (predominantly exceeding 90\% and often approaching 100\%), validates the robustness and scalability of our approach in creating separable and effective trigger-to-target mappings.

Correspondingly, the Clean Accuracy (ACC) analysis reveals that the backdoor implantation has minimal impact on the model's primary functionality. The ACC values remain stable across different trigger designs for each model-dataset combination, demonstrating that the embedded backdoors do not compromise the model's utility on clean samples. The models' maintained accuracy on clean inputs, coupled with their high vulnerability to triggered samples, highlights a clear performance disparity and underscores the stealthiness of STONE---the backdoored models preserve their expected performance on benign inputs while being susceptible to targeted manipulation when exposed to the trigger patterns.

\subsubsection{Effectiveness with Different $N$ Values}

\begin{table*}[!h]
	\centering
	\caption{One-to-N Attack Scalability with Multiple Target Classes on PointNet (Poisoning Ratio: 1\% per Target)}
	\label{tab:target_numbers_impact}
	\scriptsize 
	\setlength{\tabcolsep}{5pt} 
	\setlength{\tabcolsep}{5pt}
	\begin{tabular}{@{}llcccccccccc@{}}
		\toprule
		& & \textbf{N=0} & \multicolumn{2}{c}{\textbf{N=1}} & \multicolumn{2}{c}{\textbf{N=2}} & \multicolumn{2}{c}{\textbf{N=3}} & \multicolumn{2}{c}{\textbf{N=4}} \\
		\cmidrule(lr){4-5} \cmidrule(lr){6-7} \cmidrule(lr){8-9} \cmidrule(l){10-11}
		\textbf{Design} & \textbf{Dataset} & \textbf{ACC (\%)} & \textbf{$\overline{\textbf{ASR}}$} & \textbf{ACC} & \textbf{$\overline{\textbf{ASR}}$} & \textbf{ACC} & \textbf{$\overline{\textbf{ASR}}$} & \textbf{ACC} & \textbf{$\overline{\textbf{ASR}}$} & \textbf{ACC} \\
		& & & (\%) & (\%) & (\%) & (\%) & (\%) & (\%) & (\%) & (\%) \\
		\midrule
		\multirow{3}{*}{\makecell[c]{Single \\ Sphere}} 
		& ModelNet40 & 87.9 & 100.0 & 87.6 & 99.8 & 87.7 & 99.5 & 88.1 & 99.4 & 87.8 \\
		& ModelNet10 & 93.5 & 99.9 & 93.6 & 100.0 & 93.1 & 99.8 & 93.3 & 99.8 & 92.7 \\
		& ShapeNetPart & 98.5 & 100.0 & 98.5 & 99.9 & 98.5 & 99.9 & 98.5 & 99.9 & 98.3 \\
		\midrule
		\multirow{3}{*}{\makecell[c]{Dual \\ Sphere}} 
		& ModelNet40 & 87.9 & 100.0 & 87.6 & 99.9 & 87.7 & 99.6 & 88.1 & 99.9 & 87.9 \\
		& ModelNet10 & 93.5 & 100.0 & 93.6 & 100.0 & 93.1 & 99.9 & 93.3 & 99.8 & 93.8 \\
		& ShapeNetPart & 98.5 & 100.0 & 98.5 & 100.0 & 98.5 & 100.0 & 98.5 & 100.0 & 98.5 \\
		\bottomrule
	\end{tabular}
\end{table*}

To comprehensively evaluate the scalability of the STONE framework, we conduct experiments on the PointNet model with varying numbers of target classes ($N=1,2,3,4$) while maintaining a consistent poisoning ratio of 0.01 per target across all configurations. This systematic investigation, summarized in Table~\ref{tab:target_numbers_impact}, aims to understand how the STONE framework's performance scales with increasing complexity of multi-target backdoor implantation while examining the relationship between the total poisoning ratio, attack success rate, and model utility.

The experimental results reveal two key observations. First, the ASR remains consistently high across all target configurations, achieving near-perfect success rates (predominantly above 99\%) regardless of the number of target classes. This consistency can be attributed to the fixed per-target poisoning ratio of 0.01, which ensures sufficient poisoned samples for effective backdoor implantation for each target class. Notably, this high ASR performance is maintained for both single-sphere and dual-sphere trigger designs, demonstrating the robustness of the attack methodology. 

Second, and more importantly, the clean accuracy (ACC) exhibits remarkable stability across all experimental conditions, maintaining nearly identical performance levels (with variations within 1\%) as the total poisoning rate increases from 0\% to 4\% with growing N values. This consistent preservation of model utility on clean samples across all datasets and model architectures confirms that the backdoor implantation does not compromise the model's legitimate functionality, thereby satisfying the critical stealthiness requirement for practical backdoor attacks.

While the second observation holds across the scalability experiment, a closer examination reveals variations in ACC values between Table~\ref{tab:comprehensive_results} and Table~\ref{tab:target_numbers_impact}. For instance, with PointNet on ModelNet10, the single-sphere ACC is 90.9\% in the former versus 93.5\% in the latter. This apparent discrepancy is not a contradiction but stems directly from our model selection protocol: the final model is chosen to maximize the sum of ACC and average ASR, not each metric individually. To illustrate, when measuring ACC for ModelNet10 with 1\% poisoning ratio, selecting based on maximum (ACC + average ASR) yields ACC=90.9\% and ASR=95.6\%, whereas selecting based solely on maximum ACC gives ACC=93.3\% and ASR=94.2\%. This methodological choice explains the specific numerical differences while in no way undermining the core conclusion that backdoor implantation preserves the model's utility on clean inputs.

\subsubsection{Resistance to SOR}

\begin{table*}[!h]
	\centering	
	\caption{One-to-N Attack Performance Under SOR Defense on ShapeNetPart Dataset (Poisoning Ratio: 1\% per Target)}
	\label{tab:sor_shapenet}
	\scriptsize 
	\setlength{\tabcolsep}{5pt} 
	\setlength{\tabcolsep}{5pt}
	\begin{tabular}{@{}lcccccccccc@{}}
		\toprule
		\multirow{2}{*}{\textbf{Design}} & \multicolumn{2}{c}{\textbf{SOR Parameters}} & \multicolumn{2}{c}{\textbf{N=1}} & \multicolumn{2}{c}{\textbf{N=2}} & \multicolumn{2}{c}{\textbf{N=3}} & \multicolumn{2}{c}{\textbf{N=4}} \\
		\cmidrule(lr){2-3} \cmidrule(lr){4-5} \cmidrule(lr){6-7} \cmidrule(lr){8-9} \cmidrule(l){10-11}
		& \textbf{$top\_n$} & \textbf{$del\_n$} & $\overline{\textbf{ASR}}$ & \textbf{ACC} & $\overline{\textbf{ASR}}$ & \textbf{ACC} & \textbf{ASR} & \textbf{ACC} & $\overline{\textbf{ASR}}$ & \textbf{ACC} \\
		& & & \textbf{(\%)} & \textbf{(\%)} & \textbf{(\%)} & \textbf{(\%)} & \textbf{(\%)} & \textbf{(\%)} & \textbf{(\%)} & \textbf{(\%)} \\
		\midrule
		\multirow{3}{*}{\makecell[c]{Single \\ Sphere}} 
		& 20 & 10 & 15.1 & 97.7 & 17.5 & 98.3 & 23.6 & 96.3 & 20.9 & 97.0 \\
		& 15 & 10 & 19.0 & 98.3 & 20.2 & 95.7 & 21.6 & 97.8 & 18.1 & 94.5 \\
		& 15 & 8 & 100.0 & 98.6 & 99.9 & 98.5 & 100.0 & 98.6 & 99.9 & 98.6 \\
		\midrule
		\multirow{3}{*}{\makecell[c]{Dual \\ Sphere}} 
		& 20 & 10 & 99.8 & 98.7 & 100.0 & 98.7 & 99.9 & 98.6 & 100.0 & 98.6\\
		& 15 & 10 & 100.0 & 98.7 & 99.9 & 98.9 & 100.0 & 98.7 & 100.0 & 98.9 \\
		& 15 & 8 & 100.0 & 98.7 & 100.0 & 98.8 & 100.0 & 98.7 & 100.0 & 98.6 \\
		\bottomrule
	\end{tabular}
\end{table*}

Table \ref{tab:sor_shapenet} evaluates the effectiveness of Statistical Outlier Removal (SOR) defense against the STONE framework. The STONE framework employs the insertion of additional points in the form of small spheres around the point cloud periphery, which makes SOR the most relevant and effective countermeasure since it specifically targets statistical outliers. To rigorously test SOR's defensive capability under the most challenging conditions, we selected the 1\% poisoning ratio per target on the ShapeNetPart dataset. This configuration represents an extreme attack scenario where the backdoor attack achieves nearly 100\% Attack Success Rate (ASR) in the absence of any defense mechanisms. By testing SOR defense under these optimal attack conditions, we can more clearly demonstrate its effectiveness and limitations.

In the SOR defense, $top\_n$ defines the number of nearest neighbors used to compute local density metrics for each point, while $del\_n$ specifies the number of most distant points to be removed based on these density calculations. In general, larger values of both parameters enhance SOR's effectiveness by removing more points identified as statistical outliers.

According to the experimental results in Table~\ref{tab:sor_shapenet}, we observe distinct patterns in SOR's defensive capability against different trigger designs. For single-sphere triggers, when $\mathit{top\_n}$=15 and $\mathit{del\_n}$=8, SOR fails to defend against multi-target attacks as the ASR approaches 100\% across all target configurations. In contrast, for dual-sphere triggers, even with more stringent parameters ($\mathit{top\_n}$=20, $\mathit{del\_n}$=10), SOR still cannot effectively mitigate the attacks, maintaining near-perfect ASR. This indicates that defending against dual-sphere attacks requires further increasing $\mathit{top\_n}$ or $\mathit{del\_n}$ beyond the tested ranges. The enhanced resistance of dual-sphere triggers primarily stems from their implantation of twice as many points as single-sphere triggers, though this comes at the cost of reduced visual stealth due to the larger trigger footprint.

\section{Conclusion}
\label{sec:conclusion}
This work establishes \emph{STONE}, the first approach that realizes the one-to-N universal backdoor attack in 3D point clouds under the black box and dirty label setting---a critical threat model previously unexplored. The feasibility of this attack is supported by a key theoretical insight: the interference between distinct target mappings exhibits an exponential decay with respect to the distance between their corresponding trigger configurations in the parameter space. This property allows for the design of multiple, minimally interfering triggers within a confined spatial domain. This theoretical foundation is extensively validated through rigorous experiments across multiple datasets and network architectures, confirming that multi-target backdoor attacks represent a severe and practically viable threat to 3D vision systems.

Our work thus bridges the critical research gap in one-to-N backdoor universal attacks for 3D point clouds, transforming this concept from an intuitive possibility into a practically realized and theoretically substantiated threat. By proving the viability of parametric configuration as a robust discriminative mechanism, the STONE framework opens a new research direction and paves the way for developing more flexible, stealthy, and defense-resistant backdoor strategies in 3D domains.

\appendix  

\section{Theoretical Analysis of Spatial Specificity and Spatial Sensitivity}
\label{app:theoretical_analysis}  

\subsection{Key Assumptions and Simplifications}

The theoretical analysis relies on the following assumptions:

(1)\emph{Kernel Approximation and Similarity Modeling}: We approximate the true NTK using an RBF kernel $K(\bm{X}, \bm{X}') = \exp\left(-\gamma \|\Phi(\bm{X}) - \Phi(\bm{X}')\|^2\right)$, where $\Phi: \mathbb{R}^{K \times 3} \rightarrow \mathbb{R}^d$ captures essential geometric properties.

(2)\emph{Uniform Class Distribution}: We assume approximately uniform class distribution in benign training data, enabling the simplification $\sum_{j=1}^{H-M}K(\bm{X}, \bm{X}j)\cdot\mathbb{I}(y_j=t_i) \approx \sum{j=1}^{(H-M)/C}K(\bm{X}, \bm{X}_j)$, where $\mathbb{I}(y_j=t_i)$ is an indicator function that equals 1 if sample $j$ belongs to class $t_i$ and 0 otherwise.

(3)\emph{Small Trigger Assumption}: The spherical trigger $S_n$ is sufficiently small relative to the overall point cloud.

(4)\emph{Local Linearity of Feature Mapping}: The feature mapping $\Phi$ is locally Lipschitz continuous around $R_0$, satisfying $\|\Phi(\bm{X}'(R)) - \Phi(\bm{X}'(R_0))\| \leq L \cdot \|R - R_0\| + \mathcal{O}(\|R - R_0\|^2)$.

\subsection{Detailed Proofs}

\subsubsection{Proof of Lemma \ref{lemma:spatial_specificity}}

We analyze the poisoned classifier $f_{\bm{\theta}}(\cdot)$ through kernel regression under the Neural Tangent Kernel theory. The predictive probability for target class $t_i$ is:

\begin{equation}
	\varphi_{t_i}(\bm{X})=\frac{\sum_{j=1}^{H-M}K(\bm{X}, \bm{X}_j)\cdot\mathbb{I}(y_j=t_i)+\sum_{j=1}^{M}K(\bm{X},\bm{X}^{\prime}_j)\cdot\mathbb{I}(t_j=t_i)}{\sum_{j=1}^{H-M}K(\bm{X},\bm{X}_j)+\sum_{j=1}^{M}K(\bm{X}, \bm{X}^{\prime}_j)}
\end{equation}

Assuming uniform class distribution, we simplify to:
\begin{equation}
	\varphi_{t_i}(\bm{X})=\frac{\sum_{j=1}^{(H-M)/C}K(\bm{X},\bm{X}_j)+\sum_{j=1}^{M}K(\bm{X},\bm{X}^{\prime}_j)}{\sum_{j=1}^{H-M}K(\bm{X},\bm{X}_j)+\sum_{j=1}^{M}K(\bm{X},\bm{X}^{\prime}_j)}
\end{equation}

We define three key similarity measures:

(1) Similarity to target-class benign samples: $A(\bm{X}) = \sum_{j=1}^{(H-M)/C}K(\bm{X},\bm{X}_j)$ 

(2) Similarity to poisoned samples: $B(\bm{X}) = \sum_{j=1}^{M}K(\bm{X},\bm{X}^{\prime}_j)$ 

(3) Similarity to all benign samples: $C(\bm{X}) = \sum_{j=1}^{H-M}K(\bm{X},\bm{X}_j)$

Substituting yields:
\begin{equation}
	\varphi_{t_i}(\bm{X}) = \frac{A(\bm{X}) + B(\bm{X})}{C(\bm{X}) + B(\bm{X})} = 1 - \frac{C(\bm{X}) - A(\bm{X})}{C(\bm{X}) + B(\bm{X})}
\end{equation}

Consider $\bm{X}'_{R_0}$ (trigger at $R_0$) and $\bm{X}'_{R_1}$ (identical trigger at non-overlapping $R_1$). Due to the spatial non-overlap and the local sensitivity of the feature mapping $\Phi$, the representations differ significantly: $\Phi(\bm{X}'_{R_0})$ is not equal to $\Phi(\bm{X}'_{R_1})$, leading to reduced similarity to poisoned samples: $K(\bm{X}'_{R_1}, \bm{X}'_j) < K(\bm{X}'_{R_0}, \bm{X}'_j)$.

We quantify this similarity reduction with parameter $0 < \epsilon < 1$:
\begin{equation}
	B(\bm{X}'_{R_1}) \approx \epsilon B(\bm{X}'_{R_0})
\end{equation}

Crucially, for the similarity to benign samples, we leverage the \emph{background invariance} property: since the spherical trigger is small and localized, moving it to a different spatial position $R_1$ does not significantly alter the overall geometric structure of the point cloud beyond the immediate vicinity of the trigger. Therefore, the similarity to benign samples remains approximately unchanged:
\begin{equation}
	A(\bm{X}'_{R_1}) \approx A(\bm{X}'_{R_0}), \quad C(\bm{X}'_{R_1}) \approx C(\bm{X}'_{R_0})
\end{equation}

This background invariance is justified by the small trigger assumption and the local nature of the feature changes induced by trigger displacement.

For $\bm{X}'_{R_0}$:
\begin{equation}
	\varphi_{t_i}(\bm{X}'_{R_0}) = 1 - \frac{C(\bm{X}'_{R_0}) - A(\bm{X}'_{R_0})}{C(\bm{X}'_{R_0}) + B(\bm{X}'_{R_0})}
\end{equation}

For $\bm{X}'_{R_1}$, applying the background invariance and similarity reduction:
\begin{equation}
	\varphi_{t_i}(\bm{X}'_{R_1}) \approx 1 - \frac{C(\bm{X}'_{R_0}) - A(\bm{X}'_{R_0})}{C(\bm{X}'_{R_0}) + \epsilon B(\bm{X}'_{R_0})}
\end{equation}

Since $0 < \epsilon < 1$, we have $C(\bm{X}'_{R_0}) + \epsilon B(\bm{X}'_{R_0}) < C(\bm{X}'_{R_0}) + B(\bm{X}'_{R_0})$. With $C(\bm{X}'_{R_0}) - A(\bm{X}'_{R_0}) > 0$, it follows that:
\begin{equation}
	\frac{C(\bm{X}'_{R_0}) - A(\bm{X}'_{R_0})}{C(\bm{X}'_{R_0}) + \epsilon B(\bm{X}'_{R_0})} > \frac{C(\bm{X}'_{R_0}) - A(\bm{X}'_{R_0})}{C(\bm{X}'_{R_0}) + B(\bm{X}'_{R_0})}
\end{equation}

Therefore:
\begin{equation}
	\varphi_{t_i}(\bm{X}'_{R_1}) < \varphi_{t_i}(\bm{X}'_{R_0})
\end{equation}

Since $\varphi_{t_i}(\bm{X})$ approximates $f_{\bm{\theta}}(\bm{X})[t_i]$ under the NTK framework, we conclude:
\begin{equation}
	f_{\bm{\theta}}(\bm{X}'_{R_1})[t_i] < f_{\bm{\theta}}(\bm{X}'_{R_0})[t_i]
\end{equation}

This completes the proof. $\square$

\subsubsection{Proof of Lemma \ref{lemma:spatial_sensitivity}}
\textbf{Proof.} Building upon Lemma \ref{lemma:spatial_specificity}, we analyze the trend of the predictive probability with respect to the distance from the training position $R_0$.

From the RBF kernel properties and the local linearity of $\Phi$, the similarity between poisoned samples decays exponentially with distance:
\begin{equation}
	\frac{K(\bm{X}'_{R}, \bm{X}'_{R_0})}{K(\bm{X}'_{R_0}, \bm{X}'_{R_0})} \approx \exp\left(-\gamma L^2 \|R - R_0\|^2\right)
\end{equation}
where $L$ is the Lipschitz constant of $\Phi$.

This similarity decay affects the poisoned sample similarity term $B(\bm{X}'_{R})$. Since all poisoned samples for target class $t_i$ are generated with triggers at the fixed position $R_0$ during training, we have:
\begin{equation}
	B(\bm{X}'_{R}) \approx B(\bm{X}'_{R_0}) \cdot \exp\left(-\gamma L^2 \|R - R_0\|^2\right) + \Delta B(R)
\end{equation}
where $\Delta B(R)$ represents minor variations due to local geometric similarities, which is secondary as established in Lemma \ref{lemma:spatial_specificity}.

Now, consider the predictive probability under the background invariance assumption (i.e., $C(\bm{X}'_{R}) \approx C(\bm{X}'_{R_0})$ and $A(\bm{X}'_{R}) \approx A(\bm{X}'_{R_0})$ for small trigger displacements):
\begin{equation}
	\varphi_{t_i}(\bm{X}'_{R}) = 1 - \frac{C(\bm{X}'_{R}) - A(\bm{X}'_{R})}{C(\bm{X}'_{R}) + B(\bm{X}'_{R})} \approx 1 - \frac{C - A}{C + B(\bm{X}'_{R})}
\end{equation}
where $C = C(\bm{X}'_{R_0})$, $A = A(\bm{X}'_{R_0})$, $B_0 = B(\bm{X}'_{R_0})$, and $d = \|R - R_0\|$. The main trend of $\varphi_{t_i}(\bm{X}'_{R})$ is captured by the function:
\begin{equation}
	g(d) = 1 - \frac{C - A}{C + B_0 \cdot \exp\left(-\gamma L^2 d^2\right)} = \frac{A + B_0 \exp\left(-\gamma L^2 d^2\right)}{C + B_0 \exp\left(-\gamma L^2 d^2\right)}
\end{equation}

To rigorously analyze the monotonic behavior, consider the derivative of \( g(d) \):
\begin{equation}
	g'(d) = -\frac{(C-A)B_0 \cdot 2\gamma L^2 d \cdot \exp\left(-\gamma L^2 d^2\right)}{\left(C + B_0 \exp\left(-\gamma L^2 d^2\right)\right)^2}
\end{equation}
Since \( C > A > 0 \), \( B_0 > 0 \), \( \gamma > 0 \), \( L > 0 \), and \( d \geq 0 \), we have \( g'(d) < 0 \) for all \( d > 0 \), confirming that \( g(d) \) is strictly decreasing for \( d > 0 \). This establishes a decreasing trend. Moreover, as \( d \to \infty \), \( g(d) \to A/C \), and the approach to this baseline is governed by the exponential term, confirming an exponential decay trend.

In practice, due to local variations in \( C(\bm{X}'_{R}) \) and \( A(\bm{X}'_{R}) \), the predictive probability may not be strictly monotonic. However, the exponential decay trend dominates due to the primary effect of the similarity decay in \( B(\bm{X}'_{R}) \), as established by the RBF kernel properties and local linearity assumptions. Since \( f_{\bm{\theta}}(\bm{X}'_{R})[t_i] \approx \varphi_{t_i}(\bm{X}'_{R}) \) under the NTK framework, the model's predictive probability exhibits an exponential decay trend with respect to \( \|R - R_0\| \). \qed

\section{Trigger Placement Algorithms}
\label{app:algorithms}

\begin{algorithm}
	\caption{Single-Sphere Trigger Placement Algorithm}
	\label{alg:single_sphere}
	\begin{algorithmic}[1]
		\Require Number of targets $N$, candidate set $C_a$
		\Ensure Set of trigger positions $S_e$, where each position is a 3D point
		\State $S_e \gets \{(0.95, 0.95, 0.95)\}$ \Comment{Initialize with fixed point}
		\For{$i \gets 2$ to $N$}
		\State $\text{best\_point} \gets \emptyset$, $\text{max\_min\_dist} \gets 0$
		\For{each $p \in C_a \setminus S$} \Comment{Iterate over candidate points not in $S$}
		\State $\text{min\_dist} \gets \min_{s \in S} \|p - s\|$ \Comment{Find minimum distance to existing points}
		\If{$\text{min\_dist} > \text{max\_min\_dist}$}
		\State $\text{best\_point} \gets p$
		\State $\text{max\_min\_dist} \gets \text{min\_dist}$
		\EndIf
		\EndFor
		\State $S_e \gets S_e \cup \{\text{best\_point}\}$ \Comment{Add the best point to the set}
		\EndFor
		\State \Return $S_e$
	\end{algorithmic}
\end{algorithm}

\begin{algorithm}
	\caption{Dual-Sphere Trigger Placement Algorithm}
	\label{alg:dual_sphere}
	\begin{algorithmic}[1]
		\Require Number of targets $N$, candidate set $C_{2d}$, z-coordinate options $Z = \{0.05, 0.5, 0.95\}$
		\Ensure Set of trigger pairs $P$, where each pair consists of two points with the same x and y coordinates, and different z coordinates from $Z$
		\State $T \gets \{(0.95, 0.95)\}$ \Comment{Initialize with first fixed point}
		\For{$i \gets 2$ to $N$}
		\State $\text{best\_point} \gets \emptyset$, $\text{max\_min\_dist} \gets 0$
		\For{each $q \in C_{2d} \setminus T$} \Comment{Iterate over candidate points not in $T$}
		\State $\text{min\_dist} \gets \min_{\tau \in T} \|q - \tau\|$ \Comment{Find minimum distance to existing points}
		\If{$\text{min\_dist} > \text{max\_min\_dist}$}
		\State $\text{best\_point} \gets q$
		\State $\text{max\_min\_dist} \gets \text{min\_dist}$
		\EndIf
		\EndFor
		\State $T \gets T \cup \{\text{best\_point}\}$ \Comment{Add the best point to the set}
		\EndFor
		\State $P \gets \emptyset$
		\For{each $(x,y) \in T$}
		\State $z_1, z_2 \gets \text{SelectZPair}(Z)$ \Comment{Select two distinct z-coordinates}
		\State $P \gets P \cup \{[(x,y,z_1), (x,y,z_2)]\}$ \Comment{Convert to 3D trigger pairs}
		\EndFor
		\State \Return $P$
		\Function{SelectZPair}{$Z$}
		\State $z_1 \gets \text{RandomChoice}(Z)$
		\State $z_2 \gets \text{RandomChoice}(Z \setminus \{z_1\})$ \Comment{Select a different z-coordinate}
		\State \Return $z_1, z_2$
		\EndFunction
	\end{algorithmic}
\end{algorithm}

{}	
\end{document}